\documentclass[runningheads]{config/llncs}
 
\usepackage{config/eccv}

\usepackage{config/eccvabbrv}

\usepackage{graphicx}
\graphicspath{{images/}{tex_FIG/}}
\usepackage{booktabs}
\usepackage{svg}

\usepackage[accsupp]{axessibility}  

\usepackage{hyperref}

\usepackage{orcidlink}
\usepackage{float}
\usepackage{graphicx}
\usepackage{array}      
\usepackage{multirow}   
\usepackage{booktabs}   
\newcolumntype{C}[1]{>{\centering\arraybackslash}m{#1}}   
\usepackage{tabularx}

\usepackage{makecell}
\usepackage{wrapfig}
\usepackage{paralist}

\usepackage[dvipsnames, table]{xcolor}

\hypersetup{pdfborder={0 0 0}}

\usepackage{indentfirst} 

\usepackage[utf8]{inputenc}
\usepackage{newunicodechar}
\newunicodechar{≈}{\ensuremath{\approx}}

\definecolor{myred}{RGB}{255, 200, 200}     
\definecolor{myorange}{RGB}{255, 225, 180}  
\definecolor{myyellow}{RGB}{255, 255, 200}   

\usepackage{booktabs} 
\usepackage{array}    
\usepackage{graphicx} 
\usepackage{textcomp} 

\newcolumntype{C}[1]{>{\centering\arraybackslash}p{#1}}
\newcolumntype{L}[1]{>{\raggedright\arraybackslash}p{#1}}

\begin{document}

\title{HOIGS: Human-Object Interaction Gaussian Splatting from Monocular Videos}

\titlerunning{HOIGS: Human-Object Interaction Gaussian Splatting}

\author{Taewoo Kim\inst{1}\thanks{Both authors contributed equally.}\orcidlink{0009-0007-5389-6663} \and
Suwoong Yeom\inst{1}$^*$\orcidlink{0009-0007-4684-4785} \and
Jaehyun Pyun\inst{1}$^*$\orcidlink{0009-0004-0512-9425} \and
Geonho Cha\inst{2} \and
Dongyoon Wee\inst{2} \and
Joonsik Nam\inst{1}\orcidlink{0009-0001-7035-5603} \and
Yun-Seong Jeong\inst{1}\orcidlink{0009-0000-6737-4590} \and
Kyeongbo Kong\inst{3}\thanks{Both authors are the co-corresponding authors.} \and
Suk-Ju Kang\inst{1}$^{**}$}

\authorrunning{T. Kim et al.}

\institute{Department of Electronics Engineering, Sogang University, South Korea\\
\email{\{taewookim, suwoong96, jaehyun99, js.nam, jus8077, sjkang\}@sogang.ac.kr} \and
NAVER Cloud Corp, South Korea\\
\email{\{geonho.cha, dongyoon.wee\}@navercorp.com} \and
Department of Electrical \& Electronics Engineering, Pusan National University, South Korea\\
\email{kbkong@pusan.ac.kr}}

\maketitle

\begin{abstract}

Reconstructing dynamic scenes with complex human–object interactions is a fundamental challenge in computer vision and graphics. Existing Gaussian Splatting methods either rely on human pose priors while neglecting dynamic objects, or approximate all motions within a single field, limiting their ability to capture interaction-rich dynamics. To address this gap, we propose \textit{Human-Object Interaction Gaussian Splatting (HOIGS)}, which explicitly models interaction-induced deformation between humans and objects through a cross-attention-based HOI module. Distinct deformation baselines are employed to extract features: HexPlane for humans and Cubic Hermite Spline (CHS) for objects. By integrating these heterogeneous features, HOIGS effectively captures interdependent motions and improves deformation estimation in scenarios involving occlusion, contact, and object manipulation. Comprehensive experiments on multiple datasets demonstrate that our method consistently outperforms state-of-the-art human-centric and 4D Gaussian approaches, highlighting the importance of explicitly modeling human-object interactions for high-fidelity reconstruction. \textit{Video results are provided in the supplementary material.}




\end{abstract}

\input{tex_FIG/intro_fig_explain}
\section{Introduction}

Reconstructing dynamic scenes in Human-Object Interactions (HOI) from a monocular video is a fundamental challenge in computer vision and graphics.
Achieving high-quality view synthesis and 4D reconstruction in these scenarios is critical for downstream applications such as immersive virtual reality, metaverse content creation, and 3D animation. 
However, accurately modeling the complex spatial and temporal dependencies between humans and objects remains a challenging task, particularly given the inherent ambiguity of monocular camera inputs. 
Existing methods often struggle to jointly model the articulated motion of humans with the rigid or semi-rigid dynamics of manipulated objects, leading to reconstructions that lack physical plausibility.

Recent advancements in human-centric scene reconstruction~\cite{kocabas2024hugs,moon2024expressive,hu2024gauhuman,qian20243dgs,liu2024animatable,hu2024gaussianavatar,wen2024gomavatar,kim2025showmak3r} have successfully combined parametric human models (\textit{e.g.}, SMPL~\cite{loper2023smpl}) with 3D Gaussian Splatting (3DGS)~\cite{kerbl20233d}. 
These methods typically operate by defining a canonical space based on a reference T-pose, learning 3D Gaussian attributes via feature planes or MLPs, and deforming them into the world space using Linear Blend Skinning (LBS). 
While these approaches achieve impressive results for the human body by leveraging strong pose priors, they inherently neglect the objects being interacted with. 
Consequently, dynamic objects are frequently treated as static background elements or simply omitted, resulting in severe artifacts such as ghosting or floating geometry in the reconstructed scene (Fig.~\ref{fig:teaser}). 
By failing to explicitly model the object as a separate dynamic entity, these frameworks cannot capture the mutual influence that is essential for realistic HOI synthesis.

Conversely, general dynamic scene reconstruction methods, often referred to as 4D Gaussian Splatting~\cite{wu20244d,jung2023deformable,bae2024per,yang2023real,li2024spacetime,lee2024fully}, aim to represent all moving elements within a scene. 
These methods typically employ implicit deformation fields to map a canonical space to the world coordinate system or explicitly parameterize motion trajectories. 
However, by treating humans and objects as a unified motion field without semantic distinction, they often fail to capture the high-frequency details of human articulation and the specific rigid dynamics of objects. 
Implicit deformation fields tend to over-smooth complex motions, making it difficult to stably represent long-term non-linear trajectories.
Furthermore, without explicit interaction modeling, these methods struggle with scenarios involving close contact, occlusion, and manipulation, where the motion of the object is tightly coupled to the human.

To bridge this gap, we propose \textbf{H}uman-\textbf{O}bject \textbf{I}nteraction \textbf{G}aussian \textbf{S}platting (HOIGS), a unified framework designed to reconstruct complex dynamic scenes by explicitly modeling the mutual dependencies of human and object motions. 
Unlike prior works that isolate humans or treat the scene as a unified field, HOIGS disentangles the representation of humans and objects to apply specialized deformation strategies, while simultaneously coupling them through a novel interaction mechanism.


To accurately model the complex interactions, we introduce entity-specific deformation strategies that reflect the fundamentally distinct physical and kinematic properties of humans and objects. The human body has complex motion driven by a skeletal structure, alongside local deformations near joints that are well-captured by pose-conditioned priors. In contrast, manipulated objects are predominantly rigid structures and follow continuous spatial trajectories. To reflect these differences, we adopt separate deformation models for humans and objects.

For humans, we employ a HexPlane-based~\cite{cao2023hexplane} deformation field grounded in SMPL-X priors~\cite{pavlakos2019expressive}. By utilizing temporally varying parameters alongside spatial features, we capture fine-grained soft-tissue deformations while maintaining temporal consistency. 
For objects, we propose a Cubic Hermite Spline (CHS) motion model. 
This approach embeds velocity vectors from keyframe Gaussians with learnable parameters to ensure continuous, smooth, and physically plausible motion trajectories, preventing the temporal jitter often observed in frame-wise deformation methods.

To effectively couple these representations, we link the human and object deformation models through an HOI-aware Cross-Attention Module.
This module captures the bidirectional dependencies between the human and the object at each frame. 
By attending to the dynamic features of the human (derived from the HexPlane) and the motion features of the object (derived from the CHS embedding), the module learns to predict interaction-aware corrections. 
This explicit interaction modeling significantly mitigates interaction artifacts, particularly recovering fine hand details in challenging scenarios (Fig.~\ref{fig:teaser}).
In summary, our main contributions are as follows:
\begin{itemize}
    \item We propose \textbf{HOIGS}, a novel framework that effectively captures the distinct physical properties of humans and objects through entity-aware representations: employing a HexPlane-based model for complex human articulations and Cubic Hermite Splines (CHS) for smooth object trajectories.
    \item We introduce a Cross-Attention-based HOI Module that explicitly models the mutual influence between the human and the object, enforcing consistency during complex interactions.
    \item Extensive experiments on diverse datasets demonstrate that our method significantly outperforms state-of-the-art human-centric and general 4D Gaussian approaches in reconstructing dynamic interaction scenes from a monocular video.

\end{itemize}
\section{Related Work}

\subsection{3D Human Modeling}
Research on realistic human modeling has been extensively studied in computer vision. Early parametric models~\cite{kanazawa2018end} enabled efficient estimation of human pose but struggled to capture clothing and accessories. To address this limitation, implicit function-based methods~\cite{huang2020arch, saito2019pifu, saito2020pifuhd, xiu2022icon, xiu2023econ} were proposed to recover fine details such as hair and clothing. However, these approaches often suffer from limited global consistency and rendering efficiency. Furthermore, most methods focus primarily on human geometry, paying little attention to human-object interactions.

Following the emergence of Neural Radiance Fields (NeRF)~\cite{mildenhall2021nerf}, several works have adapted this technique for human modeling~\cite{peng2021neural, jiang2022neuman, weng2022humannerf, alldieck2022photorealistic, liao2023high, guo2023vid2avatar}, achieving realistic appearances and view consistency. However, these NeRF-based approaches still suffer from high training costs and slow rendering speeds. While some attempts~\cite{fan2024hold} have introduced objects into the reconstruction pipeline, dynamic human-object interactions remain largely unexplored.

Recently, 3D Gaussian Splatting (3DGS)~\cite{kerbl20233d} has emerged as a powerful representation and has been applied to human reconstruction~\cite{kocabas2024hugs, moon2024expressive, hu2024gauhuman, liu2024animatable, hu2024gaussianavatar}.
However, most existing methods still treat objects as static. To address this gap, we propose HOIGS, a model for stable human reconstruction in dynamic scenes that explicitly captures human-object interactions.

\subsection{Dynamic Scene Modeling}
The field of dynamic scene rendering has evolved from NeRF-based approaches~\cite{park2021nerfies, park2021hypernerf, wu2022d, fridovich2023k} to more efficient 3D Gaussian Splatting frameworks. HOSNeRF~\cite{liu2023hosnerf} effectively models human-object interactions by driving human motion through skeleton-based models such as SMPL~\cite{loper2023smpl} and leveraging object state embeddings. Nevertheless, the implicit representation inherent to NeRF introduces significant computational overhead during training and rendering, hindering the reconstruction of fine details in large-scale environments.

To address these efficiency bottlenecks, 4D Gaussian Splatting (4DGS)~\cite{wu20244d, yang2023real} extends 3DGS to the temporal domain, achieving real-time rendering speeds. However, these methods face fundamental challenges. First, they typically rely on Structure-from-Motion (SfM)~\cite{schonberger2016structure} for Gaussian initialization, which assumes a static world and thus produces inaccurate point clouds for dynamic subjects. Second, the MLP-based implicit deformation fields commonly used to model motion are adequate for simple trajectories but tend to produce oversmoothed or unnatural movements in complex human-object interaction scenarios.

To overcome these limitations, we propose an explicit, spline-based motion model that effectively captures detailed temporal dynamics. Our approach achieves robust rendering in dynamic scenes involving human-object interactions while maintaining computational efficiency.



\section{Method}
\label{sec:method}


\begin{figure*}[t]
\centerline{\includegraphics[width=\textwidth]
{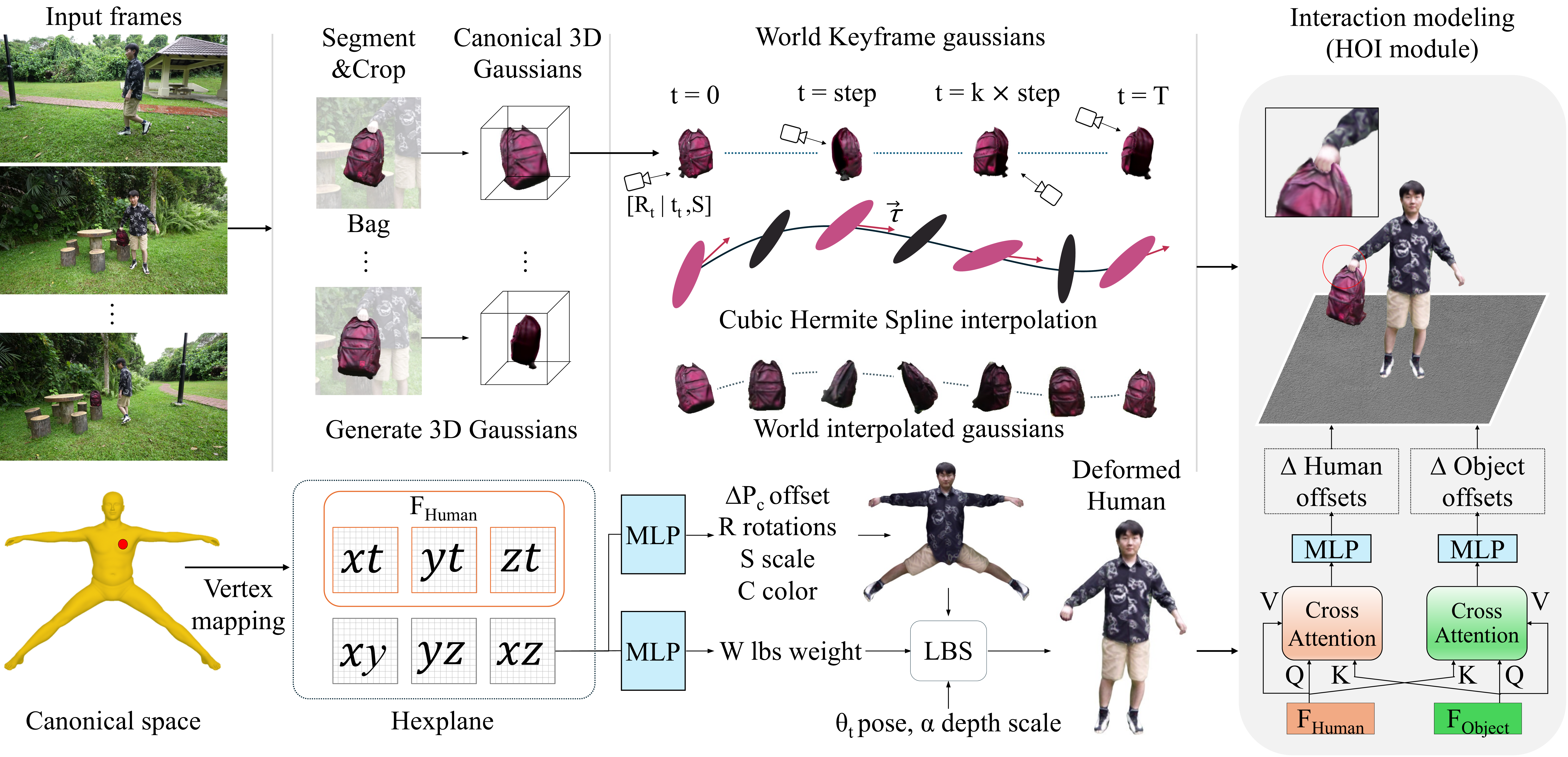}}
    \caption{\textbf{Overview of the Proposed Framework.}
    Given an input video sequence, we first extract object-specific information to reconstruct the 3D object shape via a diffusion prior. Based on the reconstructed shape, we initialize 3D Gaussians for each keyframe and use a spline-based deformation as our baseline. Here, time-invariant and time varying HexPlane features are employed for canonical human and interaction modeling, respectively. The final deformation is driven by the HOI module, which integrates HexPlane-derived human features with CHS-based object features to model human-object interactions.}
    \label{figure_main_method}
\end{figure*}

As illustrated in Fig.~\ref{figure_main_method}, our framework reconstructs dynamic scenes by disentangling the motion of humans and objects before unifying them through interaction-aware refinements. 
We model the deformation of each entity independently using distinct baselines tailored to their dynamic characteristics: a Cubic Hermite Spline (CHS) for objects (Sec.~\ref{sec:obj_deform}) and a HexPlane-based\cite{cao2023hexplane} deformation field for humans (Sec.~\ref{sec:human_deform}).
To capture the mutual influence between entities, we introduce a Cross-Attention HOI Module (Sec.~\ref{sec:hoi_module}). This module ingests motion features from both streams to regress interaction-aware geometric corrections.
Finally, the entire scene is optimized using an integrated objective function (Sec.~\ref{sec:optimization}).

\subsection{Object Deformation}
\label{sec:obj_deform}

\subsubsection{Object Initialization.}
We initialize the object representation by segmenting the target region from a representative frame and generating a canonical 3D Gaussian point cloud via a diffusion prior~\cite{dreamscene4d} guided by Score Distillation Sampling (SDS). While diffusion-based priors are capable of generating visually plausible shapes, they often suffer from scale ambiguity, failing to recover the precise scale of the actual scene geometry. 

Since these canonical Gaussians lack physical scale and world alignment, we explicitly transform them into the world coordinate system. To compute the projected 3D Gaussian bounding box area ($A_{\text{proj}}^t$), we project the 3D Gaussians onto the 2D image plane using the camera parameters at frame $t$. We then estimate a global scale factor $S$ by averaging the scale ratio between the ground-truth 2D mask bounding box area ($A_{\text{mask}}^t$) and this projected area ($A_{\text{proj}}^t$) across $T$ frames:
\begin{equation}
S = \frac{1}{T} \sum_{t=1}^{T} \sqrt{\frac{A_{\text{mask}}^t}{A_{\text{proj}}^t}}.
\end{equation}

Next, we scale the canonical Gaussian means $\boldsymbol{\mu}_{\text{can}}$ and shift them into camera space using a translation vector $\mathbf{v}_t$, which is derived from the object's masked depth. The points are then transformed into world coordinates using the camera rotation $R_t$ and translation $\mathbf{t}_t$ extracted from COLMAP~\cite{schonberger2016structure}:
\begin{equation}
\boldsymbol{\mu}_{\text{world}}^t = R_t^{-1} \big( (S \cdot \boldsymbol{\mu}_{\text{can}} + \mathbf{v}_t) - \mathbf{t}_t \big).
\end{equation}

Finally, we initialize our deformation model using these world-aligned Gaussian means $\boldsymbol{\mu}_{\text{world}}^t$ and their corresponding colors, setting the remaining Gaussian attributes (rotation, scale, and opacity) to default identity values.

\begin{figure*}[t]
\centerline{\includegraphics[width=\textwidth]
{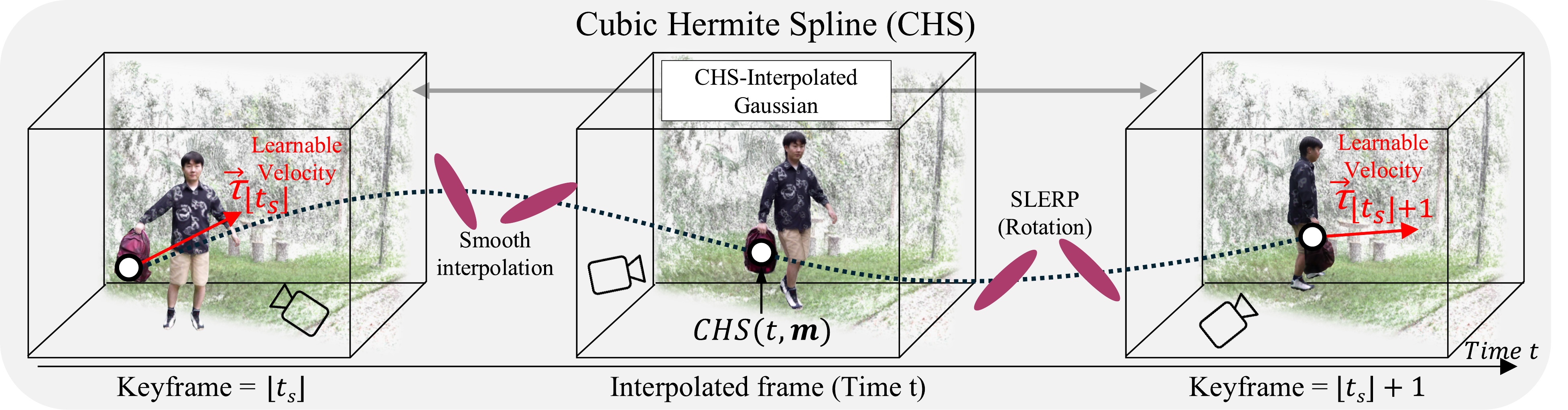}}
    \caption{Spline-based Object Motion Modeling. To ensure temporally continuous and physically plausible object trajectories, we parameterize the sequence of 3D Gaussians using a Cubic Hermite Spline (CHS).}
    \label{figure_method_sub_chs}
    \vspace{-5mm}
\end{figure*}

\subsubsection{Spline-based Motion Modeling.}
To represent continuous and smooth object motion, we model the centroids of the object Gaussians using a Cubic Hermite Spline (CHS) (Fig.~\ref{figure_method_sub_chs}).
Unlike standard discrete interpolation, CHS ensures continuity in both position and velocity, preventing abrupt jumps in the object trajectory.
We define the position of an object Gaussian at time $t$, denoted as $M(t)$, as:
\begin{equation}
M(t) = \text{CHS}(t, \mathbf{m}),
\end{equation}
where $\mathbf{m} = \{ m_k \mid m_k \in \mathbb{R}^3 \}_{k \in [0, N_{key} - 1]}$ is a learnable set of control points representing the mean positions at each keyframe, and $N_{key}$ is the number of keyframes. The spline function is formulated as:
\begin{equation}
\begin{aligned}
\text{CHS}(t, \mathbf{m}) &= (2t_r^3 - 3t_r^2 + 1)m_{\lfloor t_s \rfloor} + (t_r^3 - 2t_r^2 + t_r)\tau_{\lfloor t_s \rfloor} \\
&\quad + (-2t_r^3 + 3t_r^2)m_{\lfloor t_s \rfloor + 1} + (t_r^3 - t_r^2)\tau_{\lfloor t_s \rfloor + 1}, 
\end{aligned}
\end{equation}
where $t_r = t_s - \lfloor t_s  \rfloor $, $t_s = t_n (N_{key} - 1)$, $t_n = \frac{t}{N_f - 1}$, and $N_f$ denotes the number of all frames.
$m_{\lfloor t_s \rfloor}$ denotes the mean of the 3D Gaussians corresponding to the $\lfloor t_s \rfloor$-th key frame. We set the keyframe interval step to 4 frames during training.

Crucially, rather than approximating the tangent vector $\tau$ using the finite difference of surrounding keyframes (\textit{e.g.}, $\tau_{k} \approx \frac{1}{2}(m_{k+1} - m_{k-1})$), we reinterpret $\tau$ as an explicit \textit{velocity vector} and optimize it as a learnable parameter. This allows the model to capture complex dynamic behaviors that are not strictly inferred from positional changes alone. 
During training, gradients are backpropagated from the rendered intermediate frames to optimize both the keyframe positions $\mathbf{m}$ and velocities $\boldsymbol{\tau}$.

Regarding appearance and geometry, the rotation is modeled via Spherical Linear Interpolation (SLERP) of keyframe rotations, and opacity is time-dependent to handle occlusions. The scale parameter remains constant across keyframes to preserve object rigidity.

\subsection{Human Deformation}
\label{sec:human_deform}

We adopt a HexPlane-based representation to model human deformation, leveraging its ability to encode spatiotemporal features efficiently. 
We utilize time-invariant spatial features $f$ from the HexPlane to encode the texture of a canonical T-pose mesh $T_c$. These features are processed by an MLP head $\psi_h$ to predict Gaussian attributes in the canonical space:
\begin{equation}
\psi_h(f(T_c)) = (c, o, \Delta P_c, R, S, W),
\end{equation}
where the output tuple represents color, opacity, position offset, rotation, scale, and skinning weights, respectively.
To deform this canonical representation into the posed world space, we apply Linear Blend Skinning (LBS):
\begin{equation}
G_{human} = \alpha \cdot \text{LBS}(P_c , \theta, W),
\end{equation}
where $\theta$ represents the SMPL-X pose parameters\cite{pavlakos2019expressive} and $\alpha$ is a learnable scale factor. To ensure geometric fidelity, we enforce a depth supervision loss $\mathcal{L}_{depth} = \| D_{render} - D \|_1$, where $D$ is obtained from an off-the-shelf depth estimator\cite{video_depth_anything}, scaled via COLMAP~\cite{schonberger2016structure} points.

\subsection{Cross-Attention Interaction Module}
\label{sec:hoi_module}

\subsubsection{Feature Extraction.} We extract time-varying features from both humans and objects to learn their interactions. For humans, instead of relying on time-invariant texture features from the canonical space, we utilize time-varying features from HexPlane. Furthermore, since it is not possible to know in advance which body parts are involved in object interactions, we divide the human body into 16 parts and extract HexPlane features for each part. The human feature set is represented as
\begin{equation}
F_{\text{Human}} = \{ f_p(t) \mid p = 1, \dots, 16 \}, \quad f_p(t) = \frac{1}{|V_p|}\sum_{i \in V_p} \phi(x_i, y_i, z_i, t),
\end{equation}
where $V_p$ denotes the set of vertices in part $p$, and $\phi$ queries the HexPlane feature at time $t$. Each part feature $f_p(t) \in \mathbb{R}^{96}$, and the human feature set comprises 16 tokens, yielding a $16 \times 96$ representation corresponding to SMPL-X semantic parts (\textit{e.g.}, head, torso, arms, legs, and left/right hands).


\begin{figure*}[t]
\centerline{\includegraphics[width=\textwidth]{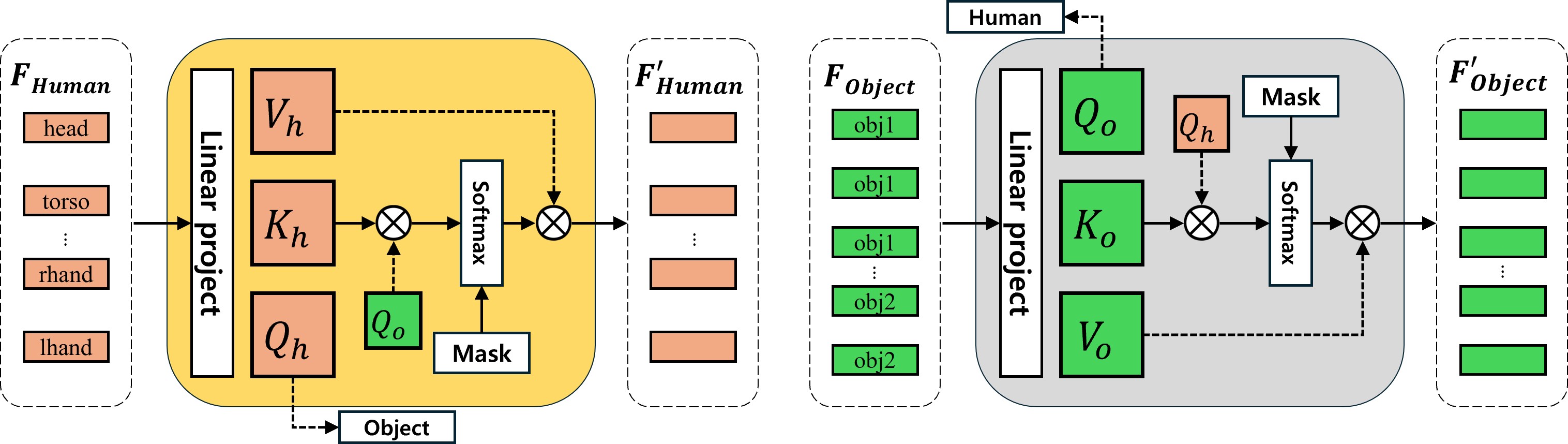}}    
    \caption{Cross-Attention-based HOI Module. The proposed architecture estimates human-object interactions using human body part features and per-Gaussian object representations.}
    \label{hoi_network}
    \vspace{-5mm}
\end{figure*}

For objects, we build features from each keyframe's velocity $\tau_k = (v_x, v_y, v_z) \in \mathbb{R}^{3}$
and a learnable embedding $e_k \in \mathbb{R}^{29}$, which are concatenated and interpolated
by CHS to yield continuous motion features. These are combined with the time value
$t \in \mathbb{R}$ and projected through a shallow MLP. Each object feature $f_o(t) \in \mathbb{R}^{32}$, and the entire object feature set comprises $N_{\text{obj}}$ such tokens, yielding an $N_{\text{obj}} \times 32$ representation:
\begin{equation}
F_{\text{Object}} = \{ f_o(t) \mid o = 1, \dots, N_{\text{obj}} \}, \quad f_o(t) = \text{MLP}\big( [\tau(t) ; e(t) ; t] \big).
\end{equation}
\subsubsection{Interaction Module.} The proposed interaction module (HOI module) takes time-varying features of the human and the object as inputs and explicitly models their interactions (Fig.~\ref{hoi_network}). We linearly project both features into 32-dimensional queries, keys, and values for \textit{mutual attention}:
\begin{equation}
\begin{aligned}
Q_h, K_h, V_h &= F_{\text{Human}}W_h^Q,\; F_{\text{Human}}W_h^K,\; F_{\text{Human}}W_h^V.
\end{aligned}
\end{equation}
\vspace{-3mm}
\begin{equation}
\begin{aligned}
Q_o, K_o, V_o &= F_{\text{Object}}W_o^Q,\; F_{\text{Object}}W_o^K,\; F_{\text{Object}}W_o^V.
\end{aligned}
\end{equation}
Cross-attention is then performed in both directions, from human to object and from object to human, while incorporating a distance mask $B$ into the attention computation:
\begin{equation}
F'_{\text{Human}} = \text{softmax}\!\left(\tfrac{Q_hK_o^\top}{\sqrt{d}} + B\right)V_h,\quad
F'_{\text{Object}} = \text{softmax}\!\left(\tfrac{Q_oK_h^\top}{\sqrt{d}} + B^\top\right)V_o.
\end{equation}

The distance mask $B$ filters out distant objects based on their 3D spatial proximity to the human.
Specifically, for each object, we compute the Euclidean distance between the object's Gaussian center $c_{\mathrm{obj}}^{\mathrm{world}}$ and the human pelvis position $p_{\mathrm{pelvis}}^{\mathrm{world}}$ in world coordinates. 
$B_{ij}$ encodes the relative distance between the $i$-th human token and the $j$-th object token:

\begin{equation}
B_{ij} = \begin{cases}
-\infty & \text{if } \lVert c_{\mathrm{obj}}^{\mathrm{world}} - p_{\mathrm{pelvis}}^{\mathrm{world}} \rVert \ge d_{\mathrm{th}} \\
0 & \text{otherwise}
\end{cases},
\end{equation}
where $d_{\mathrm{th}}$ is set to the human arm length (derived from the SMPL-X model). When $B_{ij}=-\infty$, the corresponding object tokens are masked out during attention, effectively excluding non-interacting background objects.

The resulting interaction-aware features $F'_{\text{Human}}$ and $F'_{\text{Object}}$ are used to parameterize residual corrections to the human and object. We regress SMPL-X pose refinements
$\Delta \theta = (\Delta \theta_{\text{body}}, \Delta \theta_{\text{lhand}}, \Delta \theta_{\text{rhand}})$ from $F'_{\text{Human}}$, where $\Delta \theta_{\text{body}} \in \mathbb{R}^{21 \times 3}$, $\Delta \theta_{\text{lhand}} \in \mathbb{R}^{15 \times 3}$, and $\Delta \theta_{\text{rhand}} \in \mathbb{R}^{15 \times 3}$, and object Gaussian mean offsets $\Delta G_{\text{object}} \in \mathbb{R}^{3}$ from $F'_{\text{Object}}$.
These residuals are added to the HexPlane+LBS human and the CHS object to produce interaction-consistent trajectories.

\begin{equation}
\Delta \theta = \text{MLP}_{\text{hum}}(F'_{\text{Human}}), \quad
\theta^{\text{final}} = \theta + \Delta \theta, \quad
\Delta G_{\text{object}} = \text{MLP}_{\text{obj}}(F'_{\text{Object}}).
\end{equation}

Using the refined pose and object offsets, the final Gaussians for both the human and the object are obtained as follows:
\begin{equation}
G_{\text{human}}^{\text{final}} = \alpha \cdot \text{LBS}(P_c, \theta^{\text{final}}, W), \quad
G_{\text{object}}^{\text{final}} =  M(t) + \Delta G_{\text{object}}.
\end{equation}


\subsection{Optimization}
\label{sec:optimization}

We employ an integrated optimization strategy that jointly refines the human deformation field, object motion splines, and static background representation. The comprehensive objective function balances these components alongside the depth supervision loss:
\begin{equation}
    \mathcal{L} = \mathcal{L}_{\text{human}} + \mathcal{L}_{\text{object}} + \mathcal{L}_{\text{scene}} + \mathcal{L}_{\text{depth}}
\end{equation}

\noindent\textbf{Human Loss.} 
Following Moon et al.~\cite{moon2024expressive}, we supervise the rendered human avatar using photometric losses on the cropped human region. We minimize the L1 distance, D-SSIM, and LPIPS~\cite{zhang2018unreasonable} between the rendered image $\hat{I}_{\text{human}}$ and the ground-truth image $I_{\text{gt}}$. The human loss incorporates this combined image loss $\mathcal{L}_{\text{img}}$, a mesh-based facial consistency term $\mathcal{L}_{\text{face}}$, and a Laplacian regularizer $\mathcal{L}_{\text{reg}}$ for body shape stability:
\begin{equation}
    \mathcal{L}_{\text{human}} = \mathcal{L}_{\text{img}} + \mathcal{L}_{\text{face}} + \mathcal{L}_{\text{reg}}
\end{equation}

\noindent\textbf{Object and Scene Losses.} 
We isolate the dynamic object and static background using pre-trained segmentation masks. To effectively mitigate rendering inconsistencies at complex occlusion boundaries, we compute L1 and D-SSIM photometric errors exclusively on their respective regions:
\begin{equation}
    \mathcal{L}_{\text{object}} = \mathcal{L}_{1}(I_{\text{gt}}, \hat{I}_{\text{obj}}) + \mathcal{L}_{\text{D-SSIM}}(I_{\text{gt}}, \hat{I}_{\text{obj}})
\end{equation}
\begin{equation}
    \mathcal{L}_{\text{scene}} = \mathcal{L}_{1}(I_{\text{gt}}, \hat{I}_{\text{render}}) + \mathcal{L}_{\text{D-SSIM}}(I_{\text{gt}}, \hat{I}_{\text{render}})
\end{equation}
Please refer to the supplementary material for more detailed definitions and configurations of the loss functions.

\section{Experiments}


\input{tex_FIG/Experiment_combined}

\subsection{Datasets}

For both datasets, following HOSNeRF~\cite{liu2023hosnerf}, we uniformly select 16 frames per sequence for testing and utilize the remainder for training.

\noindent\textbf{HOSNeRF.} A monocular dynamic-scene dataset capturing human--object interactions across 6 indoor/outdoor locations with 6 subjects. Each sequence contains 300--400 frames.

\noindent\textbf{BEHAVE~\cite{bhatnagar2022behave}.} Adapted from multi-view RGB-D to a monocular setting using a single fixed camera. We curate 8 sequences (labeled \textit{Object\_Subject}) comprising 300 uniformly sampled frames each, covering 4 indoor environments, 4 subjects, and 4 objects.

\noindent\textbf{ARCTIC~\cite{fan2023arctic}.} A hand--object interaction dataset, enabling comparisons with hand-object baselines. Because HOIGS is human-centric, we solely evaluate sequences where the full body is visible, utilizing sequences of 1 subject interacting with 4 objects ($\approx$600 frames/sequence).

\newcommand{\tablefirst}[0]{\cellcolor{myred}}
\newcommand{\tablesecond}[0]{\cellcolor{myorange}}

\begin{table*}[t]   
\centering
\small
\setlength\tabcolsep{1.0pt}
\caption{
Per-scene quantitative evaluation on the \textbf{HOSNeRF}\cite{liu2023hosnerf} dataset against baselines of our method. We color code each cell as \colorbox{myred}{\textbf{best}} and \colorbox{myorange}{\textbf{second best}.}}

\label{tab:hosnerf_comparison}
\resizebox{\textwidth}{!}{%
\begin{tabular}{C{4.3cm}|C{1.1cm}C{1.1cm}|C{1.1cm}C{1.1cm}|C{1.1cm}C{1.1cm}|C{1.1cm}C{1.1cm}|C{1.1cm}C{1.1cm}|C{1.1cm}C{1.1cm}}
\specialrule{.1em}{.05em}{.05em}
\multirow{2}{*}{Methods} &
\multicolumn{2}{c|}{\textbf{Backpack}} &
\multicolumn{2}{c|}{\textbf{Tennis}} &
\multicolumn{2}{c|}{\textbf{Suitcase}} &
\multicolumn{2}{c|}{\textbf{Playground}} &
\multicolumn{2}{c|}{\textbf{Dance}} &
\multicolumn{2}{c}{\textbf{Lounge}} \\
& PSNR$\uparrow$ & LPIPS$\downarrow$
& PSNR$\uparrow$ & LPIPS$\downarrow$
& PSNR$\uparrow$ & LPIPS$\downarrow$
& PSNR$\uparrow$ & LPIPS$\downarrow$
& PSNR$\uparrow$ & LPIPS$\downarrow$
& PSNR$\uparrow$ & LPIPS$\downarrow$ \\ \hline
K-Planes~\cite{fridovich2023k} 
& 19.05 & 0.557 
& 19.31 & 0.536 
& 18.64 & 0.602 
& 17.92 & 0.635 
& 18.17 & 0.623 
& 24.21 & 0.453 \\
D$^2$NeRF~\cite{wu2022d} 
& 20.52 & 0.608 
& 23.97 & 0.540 
& 20.99 & 0.645 
& 21.23 & 0.616 
& 19.92 & 0.647 
& 27.13 & 0.509 \\
Nerfies~\cite{park2021nerfies} 
& 19.56 & 0.559 
& 22.12 & 0.443 
& 19.01 & 0.555 
& 21.14 & 0.533 
& 19.37 & 0.524 
& 25.90 & 0.342 \\
HyperNeRF~\cite{park2021hypernerf} 
& 19.62 & 0.587 
& 21.26 & 0.510 
& 19.41 & 0.607 
& 21.67 & 0.578 
& 19.30 & 0.601 
& 27.25 & 0.332 \\
NeuMan~\cite{jiang2022neuman} 
& 21.21 & 0.478 
& 23.17 & 0.442 
& 20.84 & 0.551 
& 21.46 & 0.551 
& 21.19 & 0.490 
& 28.40 & 0.341 \\
4DGS~\cite{wu20244d} 
& 24.49 & 0.192 
& \tablesecond{26.57} & 0.162 
& 17.98 & 0.460 
& 24.34 & 0.222 
& 21.34 & 0.212 
& \tablesecond{30.50} & 0.067 \\
D3DGS~\cite{yang2024deformable} 
& 24.06 & \tablesecond{0.099}
& 25.09 & \tablesecond{0.125}
& 17.85 & 0.453 
& 23.93 & 0.141 
& 21.07 & \tablesecond{0.117}
& 26.90 & 0.072 \\
E-D3DGS~\cite{bae2024per} 
& \tablesecond{24.78} & 0.146 
& 26.53 & 0.161 
& 18.05 & 0.461 
& 24.37 & 0.206 
& 23.87 & 0.159 
& 30.04 & 0.086 \\
Ex4DGS~\cite{lee2024fully} 
& 18.07 & 0.433 
& 17.90 & 0.399 
& 15.25 & 0.557 
& 16.36 & 0.535 
& 17.08 & 0.529 
& 23.15 & 0.310 \\
ExAvatar~\cite{moon2024expressive} 
& 24.15 & 0.107 
& 23.57 & 0.160 
& 20.32 & \tablesecond{0.260}
& \tablefirst{25.30} & \tablesecond{0.129}
& \tablesecond{23.32} & 0.170 
& 29.43 & \tablesecond{0.048}\\
HOSNeRF~\cite{liu2023hosnerf} 
& 22.56 & 0.243 
& 24.15 & 0.320 
& \tablesecond{21.74} & 0.382 
& 22.67 & 0.336 
& 22.63 & 0.248 
& 27.74 & 0.227 \\
\hline
\textbf{HOIGS (Ours)} 
& \tablefirst{25.78} & \tablefirst{0.082}
& \tablefirst{27.12} & \tablefirst{0.108}
& \tablefirst{22.09} & \tablefirst{0.246}
& \tablesecond{25.23} & \tablefirst{0.103}
& \tablefirst{24.17} & \tablefirst{0.098}
& \tablefirst{30.97} & \tablefirst{0.048}\\
\specialrule{.1em}{.05em}{.05em}
\end{tabular}%
}
\vspace{-2mm}
\end{table*}

\begin{table*}[t]
\centering
\small
\setlength\tabcolsep{4pt}
\def\arraystretch{1.1}
\caption{
Per-scene quantitative evaluation on the \textbf{BEHAVE}\cite{bhatnagar2022behave} dataset against baselines.
}
\label{tab:BEHAVE_comparison}
\resizebox{\textwidth}{!}{%
\begin{tabular}{C{3.9cm}|C{1.4cm}C{1.4cm}|C{1.4cm}C{1.4cm}|C{1.4cm}C{1.4cm}|C{1.4cm}C{1.4cm}}
\specialrule{.1em}{.05em}{.05em}
\multirow{2}{*}{Scene}
  & \multicolumn{2}{c|}{4DGS~\cite{wu20244d}}
  & \multicolumn{2}{c|}{E\text{-}D3DGS~\cite{bae2024per}}
  & \multicolumn{2}{c|}{ExAvatar~\cite{moon2024expressive}}
  & \multicolumn{2}{c}{\textbf{HOIGS (Ours)}} \\
  & PSNR$\uparrow$ & LPIPS$\downarrow$
  & PSNR$\uparrow$ & LPIPS$\downarrow$
  & PSNR$\uparrow$ & LPIPS$\downarrow$
  & PSNR$\uparrow$ & LPIPS$\downarrow$ \\
\specialrule{.1em}{.05em}{.05em}
Backpack\_1
  & 21.81 & 0.076 & 19.99 & 0.086
  & \tablesecond{27.86} & \tablesecond{0.041}
  & \tablefirst{31.79} & \tablefirst{0.031} \\
Plasticcontainer\_1
  & 22.92 & 0.072 & 20.15 & 0.086
  & \tablesecond{29.96} & \tablesecond{0.042}
  & \tablefirst{33.10} & \tablefirst{0.032} \\
Plasticcontainer\_2
  & 26.37 & 0.081 & 24.75 & 0.078
  & \tablesecond{30.11} & \tablesecond{0.038}
  & \tablefirst{32.39} & \tablefirst{0.034} \\
Suitcase\_2
  & 26.66 & 0.071 & 25.85 & 0.058
  & \tablesecond{30.86} & \tablesecond{0.032}
  & \tablefirst{34.58} & \tablefirst{0.028} \\
Backpack\_3
  & 24.59 & 0.085 & 23.72 & 0.074
  & \tablesecond{26.47} & \tablesecond{0.054}
  & \tablefirst{30.17} & \tablefirst{0.044} \\
Plasticcontainer\_3
  & 24.60 & 0.087 & 23.81 & 0.070
  & \tablesecond{26.71} & \tablesecond{0.056}
  & \tablefirst{29.38} & \tablefirst{0.046} \\
Backpack\_6
  & 23.43 & 0.090 & 22.07 & 0.079
  & \tablesecond{25.78} & \tablesecond{0.038}
  & \tablefirst{29.05} & \tablefirst{0.030} \\
Trashbin\_6
  & 26.07 & 0.082 & 25.56 & 0.062
  & \tablesecond{29.81} & \tablesecond{0.029}
  & \tablefirst{31.62} & \tablefirst{0.023} \\
\specialrule{.05em}{.03em}{.03em}
\textbf{Average}
  & 24.56 & 0.082 & 22.99 & 0.079
  & \tablesecond{28.47} & \tablesecond{0.041}
  & \tablefirst{31.51} & \tablefirst{0.034} \\
\specialrule{.1em}{.05em}{.05em}
\end{tabular}%
}
\vspace{-2mm}
\end{table*}

\begin{figure*}[t]
    \centering
    \setlength{\tabcolsep}{1pt}
    \renewcommand{\arraystretch}{0.3}
    \resizebox{\textwidth}{!}{%
    \begin{tabular}{ccccc}
        \includegraphics[width=0.18\textwidth]{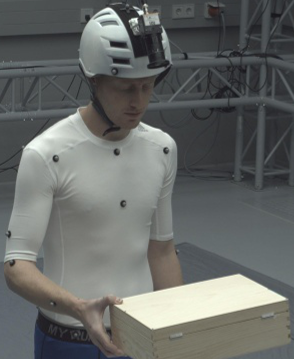} & 
        \includegraphics[width=0.18\textwidth]{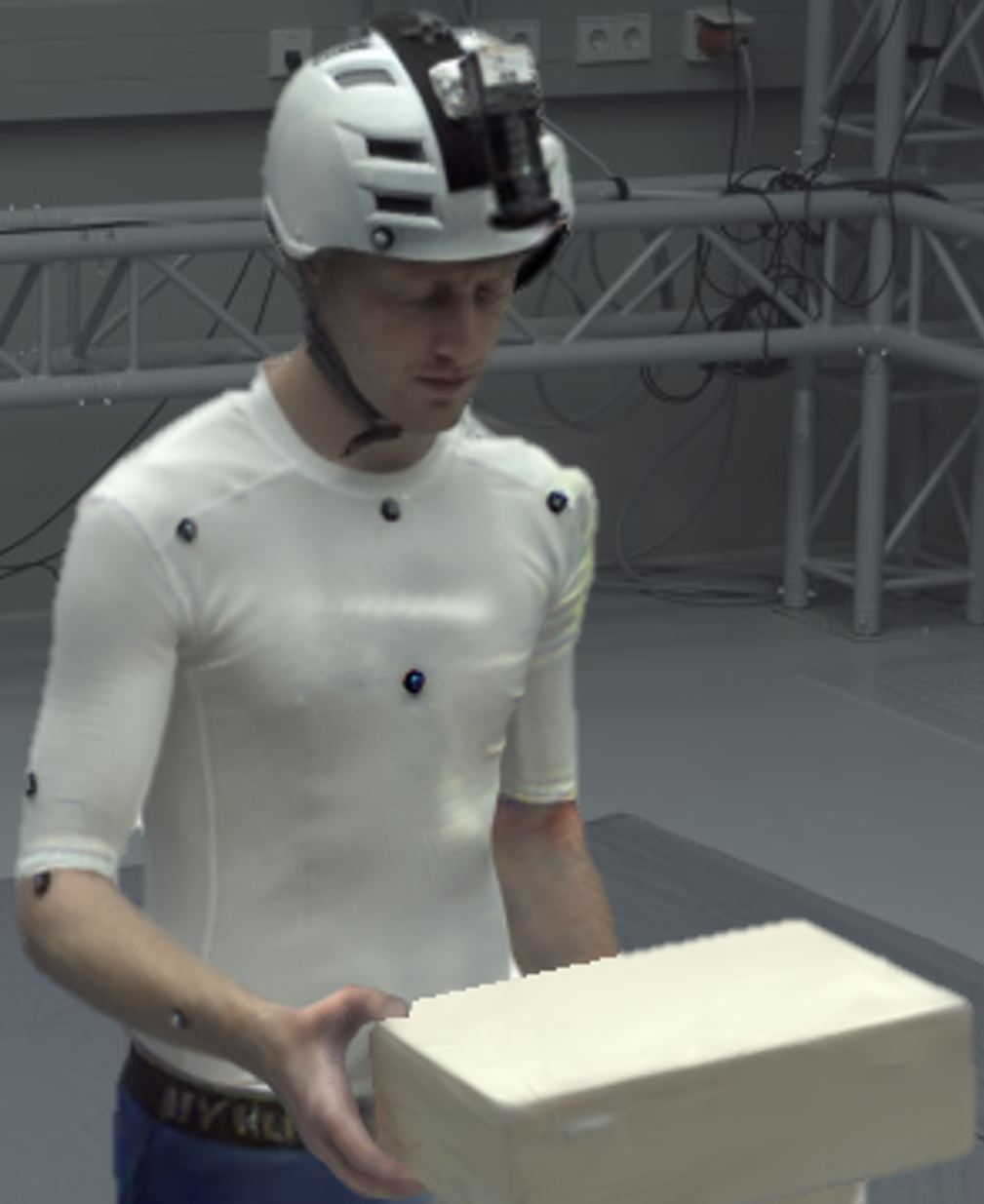} & 
        \includegraphics[width=0.18\textwidth]{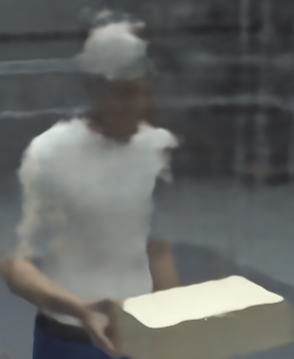} & 
        \includegraphics[width=0.18\textwidth]{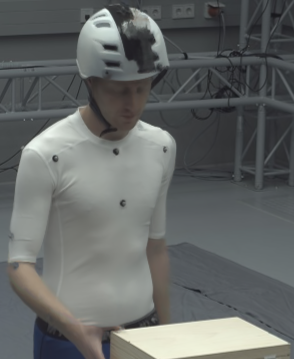} & 
        \includegraphics[width=0.18\textwidth]{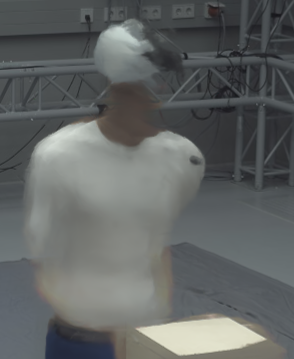} \\

        \includegraphics[width=0.18\textwidth]{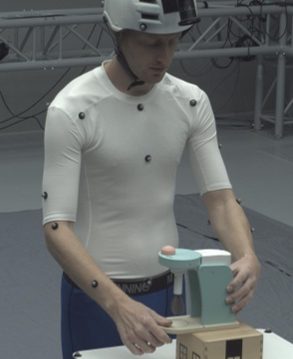} & 
        \includegraphics[width=0.18\textwidth]{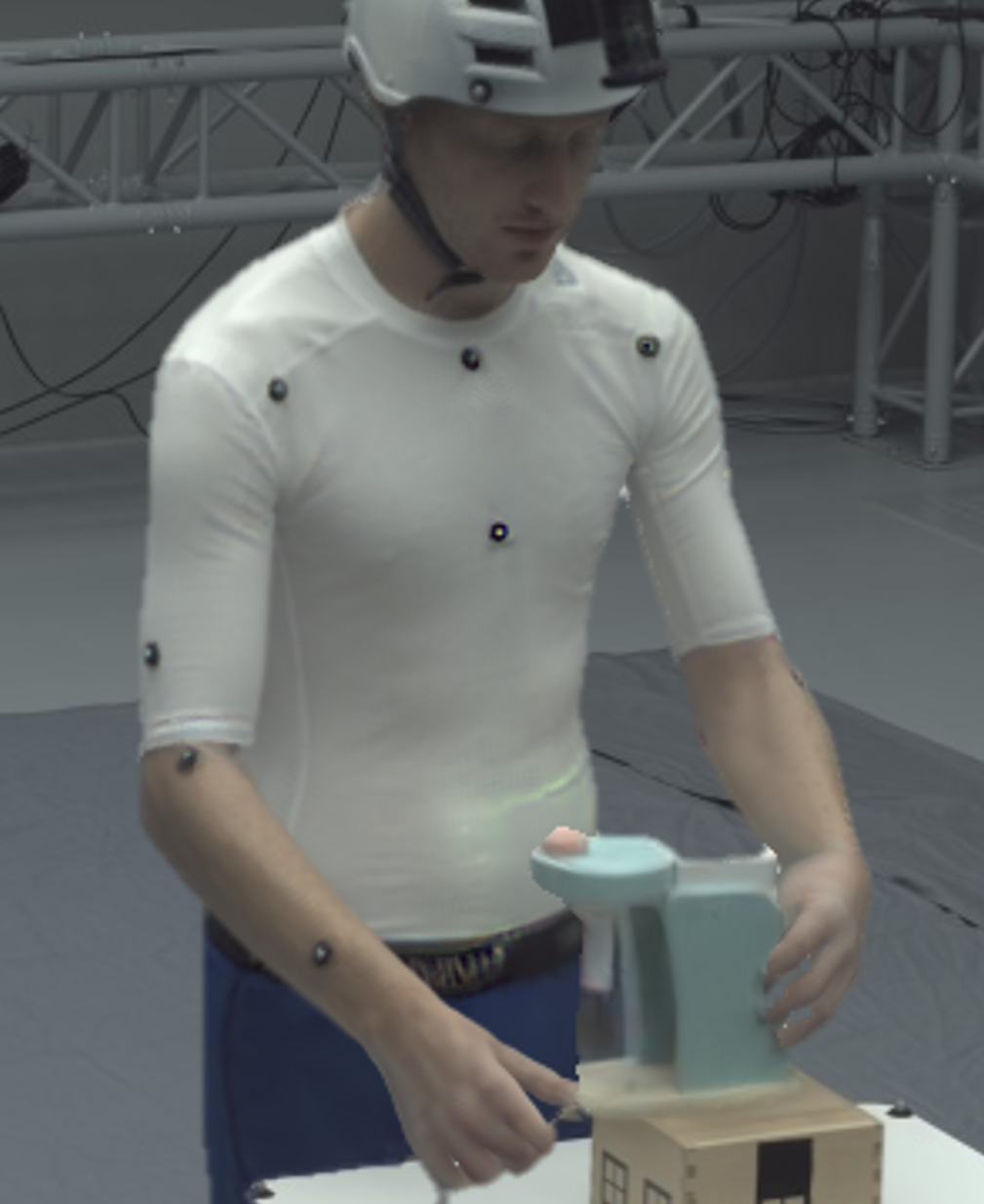} & 
        \includegraphics[width=0.18\textwidth]{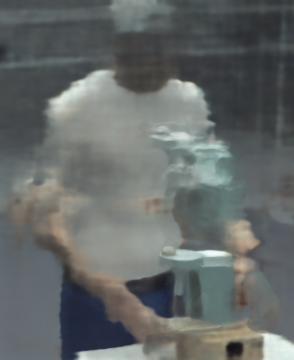} & 
        \includegraphics[width=0.18\textwidth]{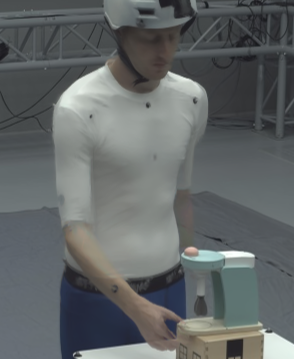} & 
        \includegraphics[width=0.18\textwidth]{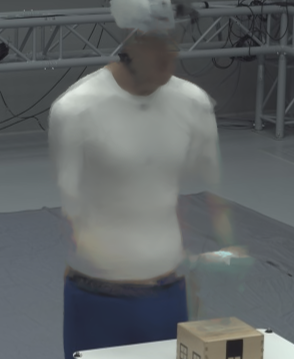} \\
        
        \footnotesize {Ground Truth} & 
        \footnotesize {Ours} & 
        \footnotesize \makecell{HOLD{~\cite{fan2024hold}}} &
        \footnotesize \makecell{E-D3DGS{~\cite{bae2024per}}} &
        \footnotesize \makecell{4DGS{~\cite{wu20244d}}} \\
        
    \end{tabular}%
    }
    \vspace{-2mm}
    \caption{Qualitative comparison of reconstructed rendered view results on the \textbf{ARCTIC}\cite{fan2023arctic} dataset.}
    \label{fig:visual_ARCTIC}
    \vspace{-2mm}
\end{figure*}

\begin{table*}[t]   
\centering
\small
\setlength\tabcolsep{1.0pt}
\def\arraystretch{1.1}
\caption{
Per-scene quantitative evaluation on the \textbf{ARCTIC}\cite{fan2023arctic} dataset against baselines of our method.
}
\label{tab:ARCTIC_comparison}
\vspace{1mm}
\resizebox{\textwidth}{!}{%
\begin{tabular}{C{4.3cm}|C{1.6cm}C{1.6cm}|C{1.6cm}C{1.6cm}|C{1.6cm}C{1.6cm}|C{1.6cm}C{1.6cm}}
\specialrule{.1em}{.05em}{.05em}
\multirow{2}{*}{Methods} &
\multicolumn{2}{c|}{\textbf{Capsulemachine}} &
\multicolumn{2}{c|}{\textbf{Mixer}} &
\multicolumn{2}{c|}{\textbf{Espressomachine}} &
\multicolumn{2}{c}{\textbf{Box}} \\
& PSNR$\uparrow$ & LPIPS$\downarrow$
& PSNR$\uparrow$ & LPIPS$\downarrow$
& PSNR$\uparrow$ & LPIPS$\downarrow$
& PSNR$\uparrow$ & LPIPS$\downarrow$ \\ \hline
4DGS~\cite{wu20244d} 
& \tablesecond{26.15} & 0.124 
& 23.21 & 0.166 
& 21.80 & \tablesecond{0.196}
& 22.22 & 0.182 \\
E\text{-}D3DGS~\cite{bae2024per} 
& 25.10 & \tablesecond{0.089}
& 22.14 & \tablesecond{0.139}
& 19.50 & 0.227 
& 20.60 & \tablesecond{0.153}\\
HOLD~\cite{fan2024hold} 
& 25.52 & 0.522 
& \tablesecond{23.35} & 0.540 
& \tablesecond{23.52} & 0.547 
& \tablefirst{24.72} & 0.494 \\
\hline
\textbf{HOIGS (Ours)} 
& \tablefirst{27.05} & \tablefirst{0.069}
& \tablefirst{24.59} & \tablefirst{0.095}
& \tablefirst{25.29} & \tablefirst{0.079}
& \tablesecond{23.50} & \tablefirst{0.124}\\
\specialrule{.1em}{.05em}{.05em}
\end{tabular}%
}
\vspace{-3mm}
\end{table*}

\subsection{Comparison with State-of-the-Art}

Tabs.~\ref{tab:hosnerf_comparison}, \ref{tab:BEHAVE_comparison}, and \ref{tab:ARCTIC_comparison} show that HOIGS consistently achieves the highest PSNR and lowest LPIPS across the HOSNeRF, BEHAVE, and ARCTIC datasets. By successfully disentangling and jointly modeling entities, our method effectively reconstructs complex human--object interactions from single-view inputs, surpassing prior 3D Gaussian dynamic scene models and human avatars. Furthermore, on the ARCTIC dataset, HOIGS outperforms the hand--object baseline HOLD~\cite{fan2024hold}. While HOLD focuses exclusively on hand reconstruction, our model reconstructs the complex full-body while simultaneously capturing dynamic object interactions.

Figs.~\ref{fig:visual_HOSNeRF} and \ref{fig:visual_BEHAVE} demonstrate that our method produces photorealistic novel views without the severe artifacts present in baseline approaches. On the HOSNeRF dataset (Fig.~\ref{fig:visual_HOSNeRF}), dynamic-scene models (D3DGS~\cite{yang2024deformable}, Ex4DGS~\cite{lee2024fully}) suffer from ghosting artifacts because they fail to disentangle human and object motions during complex interactions. ExAvatar~\cite{moon2024expressive} reconstructs the human body but completely fails to render dynamic objects. On the BEHAVE dataset (Fig.~\ref{fig:visual_BEHAVE}), ExAvatar exhibits body--background overlap due to human misalignment in world space, whereas our depth-based alignment ensures accurate spatial placement. On the ARCTIC dataset (Fig.~\ref{fig:visual_ARCTIC}), HOLD shows limited performance in full-body interactions by solely rendering hands, whereas HOIGS successfully reconstructs the entire human body alongside the interacting objects.

Fig.~\ref{fig:visual_HOSNeRF_more} shows close-ups of the interaction regions. Despite the severe occlusion and low resolution of these cropped areas, HOIGS maintains high visual fidelity, yielding sharp reconstructions that closely match the ground truth.

\begin{figure}[t]
    \centering
    \includegraphics[width=0.84\linewidth]    {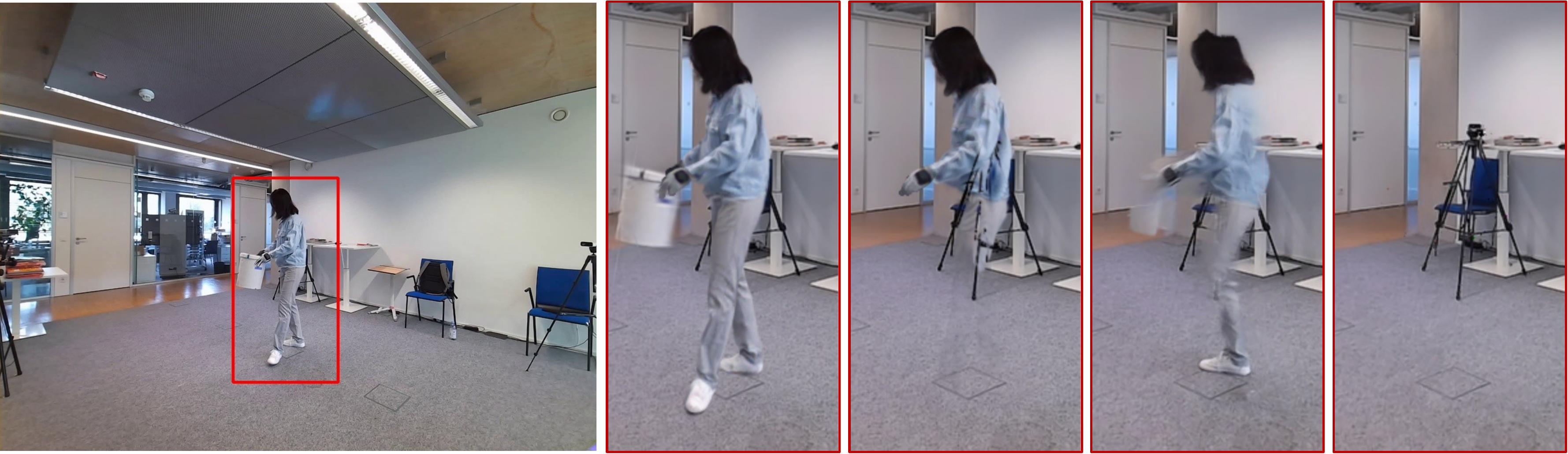}
    
    \begin{tikzpicture}[remember picture, overlay]
        \pgfmathsetmacro{\colA}{0.17}
        \pgfmathsetmacro{\colB}{0.40}
        \pgfmathsetmacro{\colC}{0.53}
        \pgfmathsetmacro{\colD}{0.66}
        \pgfmathsetmacro{\colE}{0.79}

        \def\vgap{-3mm}
        \def\leftmargin{-0.425\linewidth}
        \tikzset{label/.style={anchor=north, font=\footnotesize}}

        \node[label] at ([xshift=\leftmargin+\colA\linewidth, yshift=-\vgap] 0,0)
            {\makecell{Ground Truth}};
        \node[label] at ([xshift=\leftmargin+\colB\linewidth, yshift=-\vgap] 0,0)
            {\makecell{Ours}};
        \node[label] at ([xshift=\leftmargin+\colC\linewidth, yshift=-\vgap] 0,0)
            {\makecell{ExAvatar\\\cite{moon2024expressive}}};
        \node[label] at ([xshift=\leftmargin+\colD\linewidth, yshift=-\vgap] 0,0)
            {\makecell{E-D3DGS\\\cite{bae2024per}}};
        \node[label] at ([xshift=\leftmargin+\colE\linewidth, yshift=-\vgap] 0,0)
            {\makecell{4DGS\\\cite{wu20244d}}};
    \end{tikzpicture}
    \vspace{3mm}
    \caption{4DGS-based and human-centric models fail to render HOI scenes on the \mbox{\textbf{BEHAVE}}\cite{bhatnagar2022behave} dataset.}
    \vspace{-3mm}
    \label{fig:visual_Behave_more}
\end{figure}

Fig.~\ref{fig:visual_Behave_more} illustrates that both 4DGS and E-D3DGS completely fail to reconstruct the HOI scene. The human pose-based deformation model, ExAvatar~\cite{moon2024expressive}, fails to render the interacting object altogether. Additionally, due to inaccurate alignment in world space, ExAvatar suffers from occlusion by background Gaussians. In contrast, our model directly and accurately represents these complex human-object interactions.

\input{tex_FIG/Experiments_More_qualitative_HOSNERF}

\begin{table}[t]
\setlength{\tabcolsep}{1pt}
\begin{minipage}[t]{.48\linewidth}
\centering
\caption{Ablation study for the effectiveness of our model components on the \textbf{HOSNeRF}\cite{liu2023hosnerf} test set. The \textbf{best} results are highlighted.}
\label{ablation1}
\vspace*{-1mm}
\scalebox{0.8}{
\begin{tabular}{C{3.6cm}|C{1.2cm}C{1.2cm}}
\specialrule{.1em}{.05em}{.05em}
Settings & PSNR\textuparrow & LPIPS\textdownarrow \\ \hline
w/o CHS deformation & \multirow{2}{*}{24.52} & \multirow{2}{*}{0.154} \\
(using MLP) & & \\
Baseline deformation            & 25.01 & 0.130 \\
\textbf{HOIGS (Ours)}           & \textbf{25.89} & \textbf{0.114} \\
\specialrule{.1em}{.05em}{.05em}
\end{tabular}
}
\end{minipage}
\hspace{0.02\linewidth}
\begin{minipage}[t]{.48\linewidth}
\centering
\caption{Ablation study for the effectiveness of our interaction components on the \textbf{HOSNeRF}\cite{liu2023hosnerf} test set. The \textbf{best} results are highlighted.}
\label{ablation2}
\vspace*{-1mm}
\scalebox{0.8}{
\begin{tabular}{C{3.6cm}|C{1.2cm}C{1.2cm}}
\specialrule{.1em}{.05em}{.05em}
Settings & PSNR\textuparrow & LPIPS\textdownarrow \\ \hline
w/o Human feature               & 25.67 & 0.119 \\
w/o HOI module                  & 25.24 & 0.128 \\
\textbf{HOIGS (Ours)}           & \textbf{25.89} & \textbf{0.114} \\
\specialrule{.1em}{.05em}{.05em}
\end{tabular}
}
\end{minipage}
\end{table}



\subsection{Ablation Study}

We conduct ablation studies to validate the effectiveness of our proposed components. As shown in Tab.~\ref{ablation1}, modeling object deformation with a simple MLP yields the lowest performance, whereas our CHS-based baseline deformation improves the PSNR by 0.5, demonstrating its effectiveness. Furthermore, Tab.~\ref{ablation2} shows that removing the HOI module and relying solely on velocity results in an approximately 0.65 drop in PSNR compared to the full model, confirming the necessity of explicitly modeling human-object interactions. Finally, replacing the time-varying HexPlane features with simple parameter embeddings for the human features leads to a 0.2 decrease in PSNR, highlighting the advantage of our feature design.

\section{Conclusion}
We present \textbf{HOIGS}, a novel framework for high-fidelity monocular reconstruction of dynamic human--object interactions (HOI). By disentangling human and object motion representations, we employ a hybrid deformation strategy: HexPlane-based\cite{cao2023hexplane} deformation for detailed human articulation and Cubic Hermite Splines (CHS) for smooth, physically plausible object trajectories. Our interaction-aware HOI module couples these representations via cross-attention, capturing subtle bidirectional influences often missed in prior work. Experiments on HOSNeRF, BEHAVE, and ARCTIC \cite{liu2023hosnerf,bhatnagar2022behave,fan2023arctic} demonstrate that HOIGS significantly outperforms state-of-the-art human-centric and 4D Gaussian methods in visual quality and temporal consistency.

\noindent\textbf{Limitations.} First, we enforce object rigidity by fixing scale parameters to stabilize CHS optimization, which limits performance on highly deformable objects (\textit{e.g.}, soft tissues, and fabrics) that undergo large topological changes or compression. Second, reliance on off-the-shelf diffusion models\cite{tang2023dreamgaussian} for initial object geometry and pre-computed SMPL-X parameters\cite{pavlakos2019expressive} makes the pipeline vulnerable to initialization errors, particularly under severe occlusion.

\noindent\textbf{Future Work.} To better handle deformability, we plan to integrate physics-based priors (\textit{e.g.}, collision handling and soft-body dynamics) into the 3D Gaussian Splatting optimization. To reduce dependence on fixed preprocessing, we aim to develop a fully end-to-end pipeline that jointly refines human pose estimation and object initialization. 
\begin{center}
    \vspace*{0.5cm}
    {\Large \textsc{Supplementary Material} \textit{for}} \\[0.5cm]
    {\LARGE \textbf{HOIGS: Human-Object Interaction Gaussian Splatting \\[0.2cm] from Monocular Videos}} \\[0.5cm]
\end{center}

\setcounter{section}{0}
\setcounter{table}{0}
\setcounter{figure}{0}

\renewcommand{\thesection}{\Alph{section}}   
\renewcommand{\thetable}{\Alph{table}}   
\renewcommand{\thefigure}{\Alph{figure}}

\vspace{2mm}
\noindent\rule{\textwidth}{0.8pt}
\vspace{-2mm}
\paragraph{Overview.}
This supplementary material provides additional experiments, in-depth discussions, and further technical details omitted from the main manuscript due to space constraints. The contents are organized as follows:

\begin{itemize}
    \setlength{\itemsep}{1pt}
    \setlength{\parskip}{0pt}
    \setlength{\parsep}{0pt}
    \item \textbf{Section.~\ref{sec:arctic_evaluation}:} Geometry Evaluation of Interaction
    \item \textbf{Section.~\ref{sec:sensitivity}:} Sensitivity Analysis on External Modules
    \item \textbf{Section.~\ref{sec:complexity}:} Computational Complexity and Runtime Analysis
    \item \textbf{Section.~\ref{sec:ablation}:} Additional Ablation Studies
    \item \textbf{Section.~\ref{sec:feature}:} Feature Extraction
    \item \textbf{Section.~\ref{sec:objective}:} Objective Function Details
\end{itemize}
\vspace{-4mm}
\noindent\rule{\textwidth}{0.8pt}
\vspace{5mm}

\section{Geometry Evaluation of Interaction.}
\label{sec:arctic_evaluation}

\textbf{Motivation and Evaluation Protocol.}
Rendering metrics alone are insufficient to fully validate the geometric fidelity of human-object interactions. To comprehensively evaluate the geometric accuracy in hand-object interaction scenarios, we conducted additional evaluations on three challenging ARCTIC \cite{fan2023arctic} sequences (\textit{espressomachine\_grab\_01, mixer\_grab\_01, box\_grab\_01}) that involve complex interactions. 

While the ARCTIC sequences lack official 3D ground-truth annotations, we evaluate the geometric proximity between the reconstructed hand and object point clouds, which serves as a robust proxy for contact quality and interaction plausibility. In hand-object interaction (HOI) tasks, the fidelity of physical contact is a core objective. Therefore, measuring the inter-surface distance directly reflects how well the method resolves the interaction, making it a highly meaningful metric even in the absence of absolute ground truth. For a fair comparison, we evaluated both our method (HOIGS) and the baseline (HOLD \cite{fan2024hold}) under identical conditions, using the same input sequences and metric definitions.

\textbf{Chamfer Distance (CD).} 
To accurately measure contact quality, we compute the bidirectional Chamfer Distance between the object point cloud ($O$) and the interacting hand point cloud ($H \in \{H_L, H_R\}$). To precisely capture the interaction, we calculate the distance for both the left and right hands independently and select the minimum value, representing the hand actively engaging with the object. The distance between a single hand $H$ and the object $O$ is defined using the squared $L_2$ norm to penalize large deviations:

\begin{equation}
d_{CD}(O, H) = \frac{1}{2} \left( \frac{1}{|O|} \sum_{o \in O} \min_{h \in H} \|o - h\|_2^2 + \frac{1}{|H|} \sum_{h \in H} \min_{o \in O} \|h - o\|_2^2 \right)
\end{equation}

\begin{equation}
CD^{\mathrm{best}} = \min \left( d_{CD}(O, H_L), d_{CD}(O, H_R) \right)
\end{equation}

\begin{table}[!ht]  
\centering
\small
\renewcommand{\arraystretch}{1.2} 
\setlength\tabcolsep{12.0pt} 
\caption{
Geometric evaluation on the ARCTIC \cite{fan2023arctic} dataset. Our method (HOIGS) demonstrates significantly lower distance metrics compared to the baseline, indicating closer, more physically plausible hand-object contact. Results are averaged over all sequences.
}
\label{tab:arctic_geometry}

\resizebox{0.7\textwidth}{!}{%
\begin{tabular}{l|c|c}
\specialrule{.1em}{.05em}{.05em}
\textbf{Metric} & \textbf{HOLD \cite{fan2024hold}} & \textbf{HOIGS (Ours)} \\ \hline
$CD^{\mathrm{best}}$ [$m^2$] $\downarrow$ & 1.5035 & \textbf{0.1337} \\ 
\specialrule{.1em}{.05em}{.05em}
\end{tabular}%
}
\end{table}


As shown in Table \ref{tab:arctic_geometry}, our method significantly outperforms the baseline. HOIGS achieves a substantial improvement in Chamfer Distance compared to HOLD, demonstrating its superior ability to model physical hand-object interactions with significantly higher geometric accuracy.

\section{Sensitivity Analysis on External Modules} \label{sec:sensitivity}

\begin{table*}[t]
\centering
\small
\setlength\tabcolsep{3pt}
\caption{
Sensitivity analysis of HOIGS on the HOSNeRF\cite{liu2023hosnerf} dataset using different combinations of segmentation and depth estimation priors. The results demonstrate the robustness of our method, with consistent performance across various modern priors and strong performance even with older baselines (MaskRCNN~\cite{massa2018mrcnn}).
}
\label{tab:sensitivity_analysis}
\resizebox{\textwidth}{!}{%
\begin{tabular}{l|cc|cc|cc|cc|cc|cc|cc}
\specialrule{.1em}{.05em}{.05em}
\multirow{2}{*}{Method Combinations} &
\multicolumn{2}{c|}{\textbf{Backpack}} &
\multicolumn{2}{c|}{\textbf{Tennis}} &
\multicolumn{2}{c|}{\textbf{Suitcase}} &
\multicolumn{2}{c|}{\textbf{Playground}} &
\multicolumn{2}{c|}{\textbf{Dance}} &
\multicolumn{2}{c|}{\textbf{Lounge}} &
\multicolumn{2}{c}{\textbf{Average}} \\
& PSNR$\uparrow$ & LPIPS$\downarrow$
& PSNR$\uparrow$ & LPIPS$\downarrow$
& PSNR$\uparrow$ & LPIPS$\downarrow$
& PSNR$\uparrow$ & LPIPS$\downarrow$
& PSNR$\uparrow$ & LPIPS$\downarrow$
& PSNR$\uparrow$ & LPIPS$\downarrow$
& PSNR$\uparrow$ & LPIPS$\downarrow$ \\
\hline

Samurai~\cite{yang2024samurai} + MetricV2~\cite{hu2024metric3dv2}
& 25.78 & 0.082 & 27.12 & 0.108 & 22.09 & 0.246
& 25.23 & 0.103 & 24.17 & 0.098 & 30.97 & 0.048
& 25.89 & 0.114 \\

Samurai~\cite{yang2024samurai} + Video Depth Anything~\cite{video_depth_anything}
& 25.85 & 0.080 & 27.18 & 0.106 & 22.15 & 0.241
& 25.28 & 0.102 & 24.22 & 0.096 & 31.05 & 0.046
& 25.96 & 0.112 \\

Samurai~\cite{yang2024samurai} + DepthCrafter~\cite{hu2025-DepthCrafter}
& 25.72 & 0.088 & 27.08 & 0.108 & 22.06 & 0.249
& 25.20 & 0.109 & 24.08 & 0.099 & 30.93 & 0.048
& 25.85 & 0.117 \\

TrackAnything~\cite{yang2023track} + MetricV2~\cite{hu2024metric3dv2}
& 25.72 & 0.086 & 27.05 & 0.109 & 22.03 & 0.246
& 25.18 & 0.106 & 24.15 & 0.100 & 30.93 & 0.052
& 25.84 & 0.116 \\

SAMv2~\cite{ravi2024sam2} + Video Depth Anything~\cite{video_depth_anything}
& 26.01 & 0.076 & 27.38 & 0.103 & 22.33 & 0.241
& 25.47 & 0.100 & 24.42 & 0.095 & 31.20 & 0.041
& 26.14 & 0.109 \\

MaskRCNN~\cite{massa2018mrcnn} + MetricV2~\cite{hu2024metric3dv2}
& 25.33 & 0.099 & 26.67 & 0.125 & 21.66 & 0.257
& 24.78 & 0.117 & 23.69 & 0.110 & 30.52 & 0.065
& 25.44 & 0.129 \\

\specialrule{.1em}{.05em}{.05em}
\end{tabular}%
}
\end{table*}  

\textbf{Robustness to External Priors.}
To evaluate the impact of different external priors, we conducted a sensitivity analysis on the HOSNeRF dataset\cite{liu2023hosnerf} by evaluating our framework with various combinations of segmentation (\textit{e.g.}, Samurai~\cite{yang2024samurai}, SAMv2~\cite{ravi2024sam2}, MaskRCNN~\cite{massa2018mrcnn}, TrackAnything~\cite{yang2023track}) and depth estimation (\textit{e.g.}, Video Depth Anything~\cite{video_depth_anything}, MetricV2~\cite{hu2024metric3dv2}, DepthCrafter~\cite{hu2025-DepthCrafter}) models. 

As shown in Table~\ref{tab:sensitivity_analysis}, HOIGS maintains highly consistent performance (Avg PSNR 25.44--26.1) across different modern priors, demonstrating that our method is robust to variations in preprocessing quality. Notably, even when employing the standard, older baseline of MaskRCNN combined with MetricV2, our model achieves an average PSNR of 25.44. This performance remains significantly higher than the state-of-the-art human-centric baseline, ExAvatar \cite{moon2024expressive}(Avg PSNR 24.35), and the 4DGS baseline, Ex4DGS~\cite{lee2024fully} (Avg PSNR 17.97).

\section{Computational Complexity and Runtime Analysis} \label{sec:complexity}

\textbf{Runtime Performance.}
We evaluate the computational efficiency of our method on the HOSNeRF dataset using a single NVIDIA H100 GPU. As shown in Table~\ref{tab:runtime_comparison}, our method achieves an inference speed of \textbf{44.27 FPS}. While this is slightly lower than 4DGS~\cite{wu20244d} (61.04 FPS), it remains comparable to Ex4DGS~\cite{lee2024fully} (46.38 FPS) and outperforms D3DGS~\cite{yang2024deformable} (37.79 FPS). This result confirms that the inclusion of the HOI attention mechanism does not create a significant bottleneck, allowing our method to comfortably support real-time applications.

\begin{table}[t] 
\centering
\small
\setlength\tabcolsep{8.0pt}
\caption{
Runtime performance comparison on the HOSNeRF~\cite{liu2023hosnerf} dataset. We report the approximate training time per scene and the inference speed in Frames Per Second (FPS). Our method maintains real-time performance ($>$30 FPS) despite the added complexity of interaction modeling.
}
\label{tab:runtime_comparison}
\resizebox{0.7\textwidth}{!}{%
\begin{tabular}{l|c|c}
\specialrule{.1em}{.05em}{.05em}
\textbf{Methods} & \textbf{Training Time} & \textbf{Inference Speed (FPS)} \\ \hline
4DGS~\cite{wu20244d} & 40 min & 61.04 \\
Ex4DGS~\cite{lee2024fully} & 2 hr 30 min & 46.38 \\
D3DGS~\cite{yang2024deformable} & 3 hr & 37.79 \\
E-D3DGS~\cite{bae2024per} & 2 hr & 54.71 \\ \hline
\textbf{HOIGS (Ours)} & 5 hr & 44.27 \\
\specialrule{.1em}{.05em}{.05em}
\end{tabular}%
}
\end{table}   

\textbf{Complexity Analysis.}
The efficiency of our HOI module stems from the token-based architectural design. The cross-attention is computed between $M$ human part tokens (where $M=16$ is fixed) and $N$ object Gaussian tokens. Unlike standard self-attention which scales quadratically ($O(N^2)$), our cross-attention scales linearly ($O(M \cdot N)$) with respect to the number of object Gaussians. Furthermore, we utilize compact 33-dimensional embeddings for object motion features, which minimizes the memory footprint and matrix multiplication overhead during the forward pass.

\textbf{Discussion on Training Time.}
While our training time ($\sim$5 hours) is longer than the baselines, this is a deliberate trade-off to prioritize physical plausibility and interaction accuracy. Explicitly modeling mutual dependencies and backpropagating gradients through the attention mechanism requires more iterations. However, this cost is strictly confined to the offline training phase, ensuring that the final online user experience remains real-time.

\begin{table*}[t]
\centering
\small
\setlength\tabcolsep{1.0pt}
\caption{
Quantitative ablation study on Object Priors using the HOSNeRF~\cite{liu2023hosnerf} dataset. We evaluate the effectiveness of our Diffusion Prior against MASt3R~\cite{duisterhof2025mastrsfm} and Depth Reconstruction priors.
}
\label{tab:prior_ablation}
\resizebox{\textwidth}{!}{%
\begin{tabular}{l|C{1.1cm}C{1.1cm}|C{1.1cm}C{1.1cm}|C{1.1cm}C{1.1cm}|C{1.1cm}C{1.1cm}|C{1.1cm}C{1.1cm}|C{1.1cm}C{1.1cm}|C{1.1cm}C{1.1cm}}
\specialrule{.1em}{.05em}{.05em}
\multirow{2}{*}{Methods} &
\multicolumn{2}{c|}{\textbf{Backpack}} &
\multicolumn{2}{c|}{\textbf{Tennis}} &
\multicolumn{2}{c|}{\textbf{Suitcase}} &
\multicolumn{2}{c|}{\textbf{Playground}} &
\multicolumn{2}{c|}{\textbf{Dance}} &
\multicolumn{2}{c|}{\textbf{Lounge}} &
\multicolumn{2}{c}{\textbf{Average}} \\
& PSNR$\uparrow$ & LPIPS$\downarrow$
& PSNR$\uparrow$ & LPIPS$\downarrow$
& PSNR$\uparrow$ & LPIPS$\downarrow$
& PSNR$\uparrow$ & LPIPS$\downarrow$
& PSNR$\uparrow$ & LPIPS$\downarrow$
& PSNR$\uparrow$ & LPIPS$\downarrow$
& PSNR$\uparrow$ & LPIPS$\downarrow$ \\ \hline

MASt3R Prior
& 23.51 & 0.135
& 25.25 & 0.121
& 22.40 & 0.197
& 24.65 & 0.074
& 23.63 & 0.115
& 28.99 & 0.057
& 24.59 & 0.128 \\

Depth Recon Prior
& 21.63 & 0.142
& 25.65 & 0.122
& 22.13 & 0.230
& 25.24 & 0.103
& 24.08 & 0.123
& 28.95 & 0.095
& 25.36 & 0.136 \\

\hline
\textbf{Diffusion Prior (Ours)}
& 23.70 & 0.082
& 27.13 & 0.112
& 22.96 & 0.235
& 25.63 & 0.123
& 24.17 & 0.093
& 29.97 & 0.043
& 25.89 & 0.114 \\

\specialrule{.1em}{.05em}{.05em}
\end{tabular}%
}
\end{table*}

\section{Additional Ablation Studies} \label{sec:ablation}

\textbf{Impact of Object Diffusion Prior.}
To validate the effectiveness of our design choice, we investigate the impact of different geometric priors on the final reconstruction quality. We compare our method, which utilizes a generative Diffusion Prior, against two alternative initialization strategies:
\begin{enumerate}
    \item \textbf{MASt3R Prior:} Initialization using MASt3R~\cite{duisterhof2025mastrsfm}, a state-of-the-art dense matching and reconstruction model.
    \item \textbf{Depth Reconstruction Prior:} Initialization using standard monocular metric depth estimation.
\end{enumerate}

Table \ref{tab:prior_ablation} presents the quantitative comparison on the HOSNeRF dataset. Our method equipped with the Diffusion Prior achieves the highest average reconstruction quality (25.89 PSNR), outperforming the MASt3R prior (24.59 PSNR) and the Depth prior (25.36 PSNR). While discriminative approaches like MASt3R or metric depth estimation rely heavily on visible cues, they often struggle to reconstruct accurate geometry in the presence of heavy occlusions, a common occurrence in human-object interaction scenarios (\textit{e.g.}, hands covering objects). In contrast, the Diffusion Prior leverages generative knowledge to plausibly complete 3D geometry even in occluded or unseen regions. This holistic geometric initialization provides a more robust starting point for our Cubic Hermite Spline (CHS) deformation, leading to sharper rendering and more stable tracking throughout the dynamic sequence.

\section{Feature Extraction} \label{sec:feature}

\begin{figure*}[htb!]
\centerline{\includegraphics[width=\textwidth]{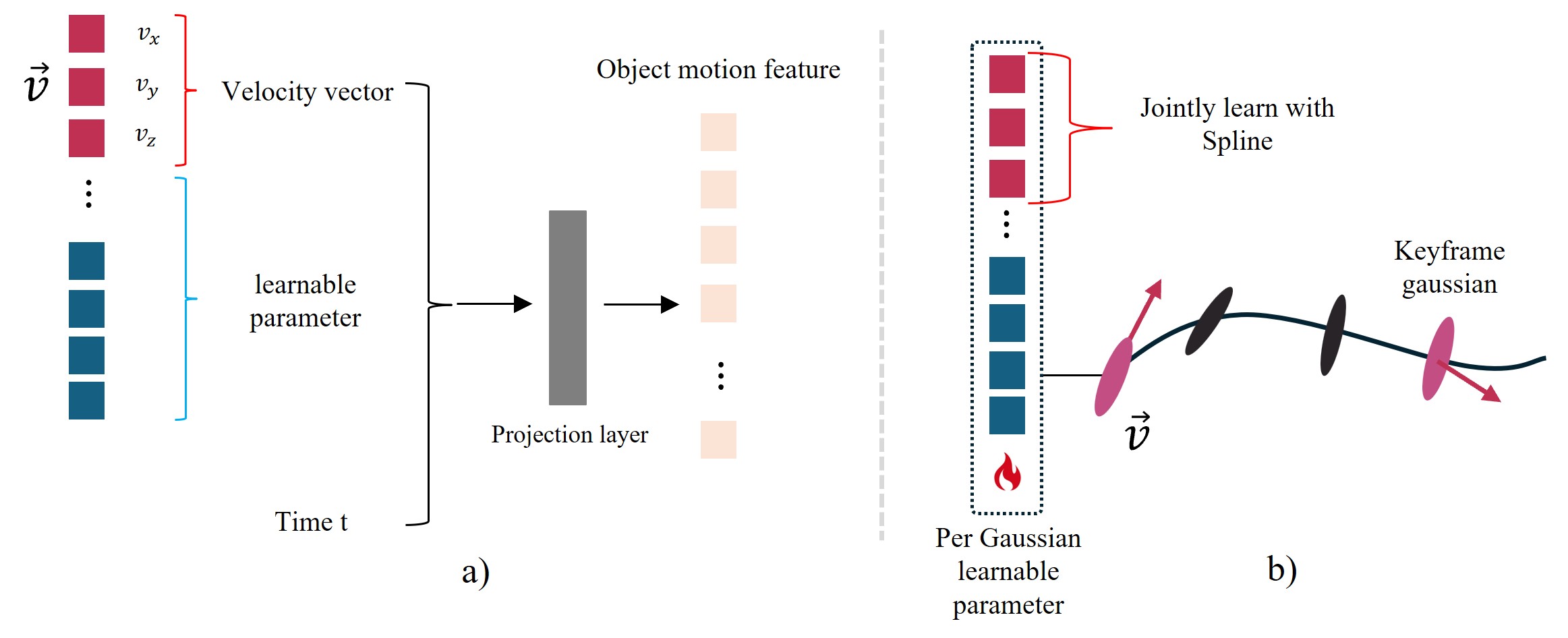}}
    \vspace{-2mm}
    \caption{\textbf{Object feature extraction.}
    Extraction of object motion features using the embedded parameters and velocity vectors of each key frame.}
    \label{appendix_objfeature}
    \vspace{-1mm}
\end{figure*}


\textbf{Object Feature.}
As shown in Fig. \ref{appendix_objfeature}(a), we extract object features by leveraging the velocity vectors and embedding parameters of Gaussians at keyframes. As illustrated in Fig. \ref{appendix_objfeature}(b), each keyframe’s velocity vector, denoted as $\tau_k = (v_x, v_y, v_z) \in \mathbb{R}^{3}$, is applied to the Cubic Hermite Spline (CHS) and jointly optimized alongside the baseline deformation to serve as input features for the HOI module. Furthermore, we introduce a 29-dimensional learnable embedding $e_k \in \mathbb{R}^{29}$ for each keyframe Gaussian, which is concatenated with the velocity vector.

To yield continuous motion features, these concatenated vectors are interpolated through the CHS. The interpolated Gaussian features are then combined with the temporal variable $t \in \mathbb{R}$ and projected through a shallow Multi-Layer Perceptron (MLP). This process results in a 33-dimensional feature vector for each object token $f_o(t) \in \mathbb{R}^{33}$. The entire object feature set comprises $N_{\text{obj}}$ such tokens, yielding a comprehensive $N_{\text{obj}} \times 33$ representation. This extraction process is formally defined as:


\begin{equation}
F_{\text{Object}} = \{ f_o(t) \mid o = 1, \dots, N_{\text{obj}} \}, \quad f_o(t) = \text{MLP}\big( [\tau(t) ; e(t) ; t] \big).
\end{equation}

\textbf{Human Feature.}
To construct the human representation, we utilize time-varying features directly from the HexPlane\cite{cao2023hexplane} rather than relying on time-invariant texture features from the canonical space. Because it is not possible to know in advance which specific body parts will be involved in object interactions, we divide the SMPL-X~\cite{pavlakos2019expressive} model into 16 distinct semantic parts and extract corresponding HexPlane features for each.

Temporal features are sampled from the HexPlane at the SMPL-X vertices, where each feature at time $t$ is obtained based on the dynamic coordinates $(x_i, y_i, z_i)$. For each body part, the features of its associated vertices are averaged to form the part-specific representation. The resulting human feature set is represented as:

\begin{equation}
F_{\text{Human}} = \{ f_p(t) \mid p = 1, \dots, 16 \}, \quad f_p(t) = \frac{1}{|V_p|}\sum_{i \in V_p} \phi(x_i, y_i, z_i, t),
\end{equation}

Here, $V_p$ denotes the specific set of vertices belonging to part $p$, and $\phi$ is the function querying the HexPlane feature at time $t$. Each resulting part feature $f_p(t)$ resides in $\mathbb{R}^{96}$. Consequently, the human feature set comprises 16 tokens, yielding a $16 \times 96$ tokenized representation that corresponds to meaningful SMPL-X semantic regions (e.g., head, torso, arms, legs, and left/right hands).

\section{Objective Function Details} \label{sec:objective}




We employ an integrated optimization strategy that jointly refines the human deformation field, object motion splines, and static background representation. To ensure consistent interaction modeling, the entire scene is optimized simultaneously using a unified objective function. The total loss $\mathcal{L}$ is defined as:
\begin{equation}
\mathcal{L} = \lambda_{\text{hum}} \mathcal{L}_{\text{human}} + \lambda_{\text{obj}} \mathcal{L}_{\text{object}} + \lambda_{\text{scene}} \mathcal{L}_{\text{scene}} + \lambda_{\text{depth}} \mathcal{L}_{\text{depth}}
\end{equation}
The weights $\lambda_{\text{hum}}$, $\lambda_{\text{obj}}$, $\lambda_{\text{scene}}$, and $\lambda_{\text{depth}}$ control the relative importance of each term and are empirically set to 0.5, 1.0, 0.25, and 1.0, respectively.

\textbf{Human Loss.} Following Moon et al. \cite{moon2024expressive}, we supervise the rendered human avatar to ensure high-fidelity appearance and structural stability. Diverging from relying solely on skeletal joint errors, our human loss formulation focuses on photometric accuracy, facial detail preservation, and topological consistency. The total human loss, $\mathcal{L}_{human}$, comprises three primary components: a combined photometric image loss ($\mathcal{L}_{img}$), a mesh-based facial consistency term ($\mathcal{L}_{face}$), and a Laplacian regularizer ($\mathcal{L}_{reg}$).

Specifically, the photometric image loss, $\mathcal{L}_{img}$, is computed exclusively on the cropped human region to isolate the subject's appearance from the background. This term is designed to minimize the visual discrepancies between the rendered human image $\hat{I}_{human}$ and the corresponding ground-truth image $I_{gt}$ by concurrently evaluating the $L_1$ distance, D-SSIM, and LPIPS~\cite{zhang2018unreasonable} metrics. To further maintain geometric and textural coherence in highly expressive areas, $\mathcal{L}_{face}$ is applied to explicitly enforce mesh-based facial consistency. Finally, the Laplacian regularizer, $\mathcal{L}_{reg}$, is applied to the underlying mesh to prevent degenerate deformations and enforce global body shape stability during the optimization process. Formally, the comprehensive human loss is formulated as:

\begin{equation}
\mathcal{L}_{img} = \mathcal{L}_{1}(\hat{I}_{human}, I_{gt}) + \mathcal{L}_{\text{D-SSIM}}(\hat{I}_{human}, I_{gt}) + \mathcal{L}_{\text{LPIPS}}(\hat{I}_{human}, I_{gt})
\end{equation}
\begin{equation}
\mathcal{L}_{human} = \mathcal{L}_{img} + \mathcal{L}_{face} + \mathcal{L}_{reg}
\end{equation}

\textbf{Object Loss.}
The $\mathcal{L}_{\text{object}}$ term is a photometric loss that focuses exclusively on the object regions within the scene. We render only the segmented object areas and compute the loss solely on these regions. A pre-trained object segmentation model is employed to isolate object masks in the input images. The object loss encourages accurate reconstruction and appearance consistency for moving objects, which often undergo significant deformation and motion. By supervising only the object regions, this loss helps to refine the geometry and texture of the object-specific Gaussians without being influenced by background or human-related elements. The object loss is defined as:

\begin{equation}
    \mathcal{L}_{\text{object}} = 0.8\,\mathcal{L}_{1}(I_{\text{gt}}, \hat{I}_{\text{obj}}) + 0.2\,\mathcal{L}_{\text{D-SSIM}}(I_{\text{gt}}, \hat{I}_{\text{obj}})
\end{equation}



\textbf{Scene Loss.}
The $L_{\text{scene}}$ term is a photometric loss focusing on the background regions of the entire scene, following the image similarity-based loss used in existing 3D Gaussian Splatting~\cite{kerbl20233d} (3DGS) methods. Specifically, a pre-trained human/object segmentation model is employed to mask out human and object regions in the images, optimizing the background Gaussians for the remaining pixels only. This involves minimizing the difference between the rendered result and the background pixels excluding the segmented human and object areas. Occlusions frequently occur during interactions between human hands and objects, causing inconsistencies in masks. By optimizing humans, objects, and backgrounds simultaneously, our method effectively mitigates these boundary inconsistencies. The scene loss is explicitly defined as:

\begin{equation}
    \mathcal{L}_{\text{scene}} = 0.8\,\mathcal{L}_{1}(I_{\text{gt}}, \hat{I}_{\text{render}}) + 0.2\,\mathcal{L}_{\text{D-SSIM}}(I_{\text{gt}}, \hat{I}_{\text{render}})
\end{equation}


\bibliographystyle{bib/splncs04}
\bibliography{bib/iclr2026_conference}
\end{document}